\documentclass[runningheads]{llncs}

\usepackage{hyperref}
\usepackage{graphicx}
\usepackage[utf8]{inputenc}

\title{Key Information Extraction From Documents: Evaluation And Generator\thanks{Supported by organization KI Group GmbH.}}

\author{Oliver Bensch\inst{1} \and
Mirela Popa\inst{2}\and
Constantin Spille\inst{3}}
\authorrunning{O. Bensch et al.}
\institute{Maastricht University, 6200 MD Maastricht, The Netherlands \and
Maastricht University, 6200 MD Maastricht, The Netherlands \and
 KI Group GmbH \\
\email{o.bensch@student.maastrichtuniversity.nl}\\ 
\email{mirela.popa@maastrichtuniversity.nl}\\
\email{c.spille@kigroup.de}\\
} 

\begin{document}

\maketitle

\begin{abstract}
Extracting information from documents usually relies on natural language processing methods working on one-dimensional sequences of text. In some cases, for example, for the extraction of key information from semi-structured documents, such as invoice-documents, spatial and formatting information of text are crucial to understand the contextual meaning. Convolutional neural networks are already common in computer vision models to process and extract relationships in multidimensional data. Therefore, natural language processing models have already been combined with computer vision models in the past, to benefit from e.g. positional information and to improve performance of these key information extraction models. Existing models were either trained on unpublished data sets or on an annotated collection of receipts, which did not focus on PDF-like documents. Hence, in this research project a template-based document generator was created to compare state-of-the-art models for information extraction. An existing information extraction model “Chargrid” (Katti et al., 2019) was reconstructed and the impact of a bounding box regression decoder, as well as the impact of an NLP pre-processing step was evaluated for information extraction from documents. The results have shown that NLP based pre-processing is beneficial for model performance. However, the use of a bounding box regression decoder increases the model performance only for fields that do not follow a rectangular shape.
\end{abstract}

\section{Introduction}
Natural language processing (NLP) methods are widely used on one-dimensional sequences of text. In some cases, for example, in the extraction of key information of invoice documents, spatial information such as the position of text are crucial to understand the contextual meaning. 

Convolutional neural networks (CNNs) are already common in computer vision (CV) models to process and extract relationships in multidimensional data. CV models have already been combined with NLP models in the past \cite{Chargrid},\cite{BERTgrid},\cite{CUTIE}, to benefit from spatial information. However, existing models were either trained on not published data sets \cite{Chargrid}, \cite{BERTgrid} or on a public available annotated collection of receipts \cite{SROIE}, which does not focus on PDF-like documents. Therefore, no benchmark dataset is currently available for comparing or evaluating existing models that extract information from documents.

These complex models consist of several components, such as an encoder, a semantic segmentation and bounding box regression that are trained on different targets with different loss functions. All this results in a substantial increase of the model's parameters and overall training time, e.g., the addition of the bounding box regression decoder as used in \cite{Chargrid} increases the trainable parameters by over 75\,\%. However, a detailed analysis of the effects of those components on overall model performance has not been published.

In this research, a template based document generator is created to generate datasets of semi-structured documents, that can be used to compare models for information extraction from documents. Furthermore, this research evaluates the possible benefits of a bounding box regression decoder as the first step to analyse the effect of the individual components. As BERTgrid has shown to improve model performance by applying a NLP pre-processing step, this research evaluates the benefits of an NLP pre-processing step also on a generated dataset of invoices. Therefore, the main contribution of this work resides in providing a comparative analysis of state-of-the-art models on a publicly available dataset \cite{DriveDataset}, \cite{GitHubCode}. The accuracy achieved on various document fields is assessed and discussed in sections \ref{Results} and \ref{Discussion}. 

\section{Related Work}
Chargrid \cite{Chargrid} proposed a model to extract information from documents using an auto-encoder with CNNs. This model has shown to outperform NLP based methods on key information extraction from documents. 

Chargrid uses the optical character recognition engine Tesseract \cite{Tesseract} to one-hot encode a documents characters per pixel with a dimension added representing the background. These pixel vectors were down-scaled and fed as an input to the model. BERTgrid \cite{BERTgrid} replaces the one-hot encoded character vectors by word embeddings, generated with  BERT \cite{BERT-Model}. Bertgrid was trained and evaluated on the same closed dataset as Chargrid and has shown to improve performance on list fields. 

Another state-of-the-art model that combines CV methods with NLP methods is CUTIE \cite{CUTIE}. CUTIE uses Tesseract to extract textual information of a document. The detected text is first mapped to a table and used as an input for their proposed CUTIE-A and CUTIE-B models. The extracted table information is compressed by an embedding layer, followed by CNN blocks to classify the input. 

\section{Method}

\subsection{Template Based Document Generator}
\begin{figure}
\centering
\includegraphics[width=0.8\textwidth]{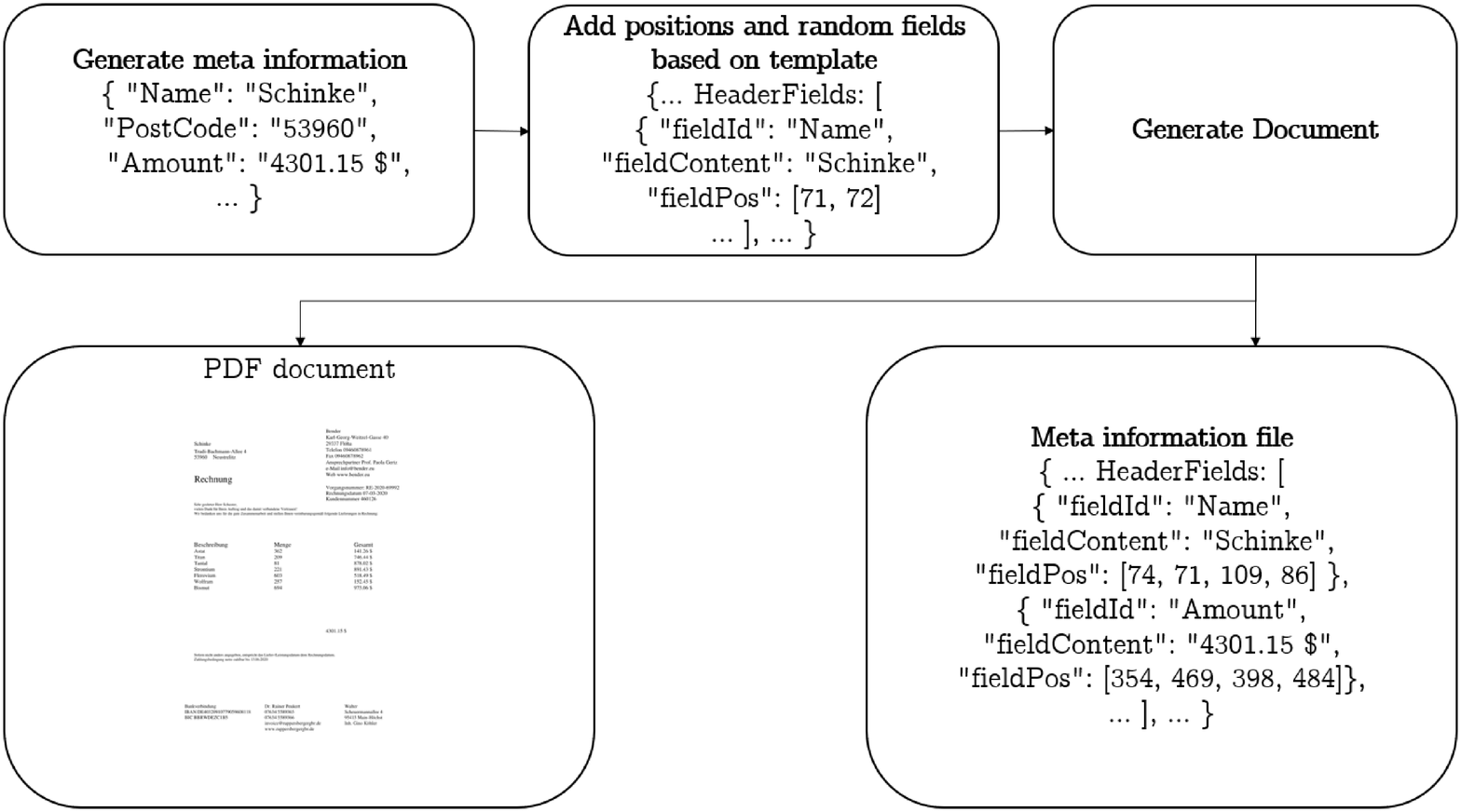}
\caption{Flowchart of the document generation process.} \label{FlowChartDocGeneration}
\end{figure}

The generation process can be seen in figure \ref{FlowChartDocGeneration}.
First, all information to be extracted from each document is generated in JSON-Format \cite{JSON}. In a next step, spatial information is added to each field, using dynamic template elements from previously manually generated templates. These templates also add fields that should be displayed but not extracted. The resulting blocks are converted into PDF-documents using PyMuPDF \cite{MuPDF}. Positional information about the fields and their labels are stored in JSON to generate data for model training.

The generated PDF-documents are available online \cite{DriveDataset}. It can be seen that the generated documents share features to real world examples. However, especially the list item fields follow more complex structures in the industry.

\subsection{Pre-Processing}
The generated PDF-Documents are pre-processed for training, as described in the Chargrid model \cite{Chargrid}. The second evaluated model SpaCygrid does not use characters as an input, rather a vector representation of the words, similar to BERTgrid \cite{BERTgrid}.

\subsection{Models} \label{Models}
In this research, the model structure presented in Chargrid \cite{Chargrid} is evaluated. To evaluate the performance trade-off between the models components, different compositions of this model are subsequently tested.

The presented model follows the auto-encoder structure with one encoder and up to three decoders which are composed of several CNN blocks. The first decoder forms the semantic segmentation decoder, while the two others form the bounding box regression decoder. All three decoders are based on the same structure and can be differentiated from each other by the last convolutional block, ending with a specific convolutional layer for each decoder. The semantic segmentation decoder uses a softmax activation function with an output dimension of $364 \times 256 \times F$, where $F$ describes the amount of fields to be extracted. The last convolutional layer of the first bounding box regression decoder, also called the box mask decoder, uses a softmax activation function as well, but with an output dimension of $W \times H \times (2N)$. Where $N$ describes the amount of anchors, with one foreground and one background layer per anchor. The second decoder of the bounding box regression decoder uses a linear activation function. The output dimensions of this decoder are $W \times H \times (4N)$, where positional as well as field size were encoded per anchor.

The full model of the Chargrid paper further follows an U-Net structure. This step increases the training parameters for the models by nearly the factor two. Due to the size of the models and the resulting training and evaluation time, the impact of this step is not yet evaluated in this research.

\subsection{Evaluation} \label{Evaluation}
The output of the semantic segmentation decoder creates prediction masks for each field that should be extracted. OpenCV \cite{OpenCV} is used to extract the bounding boxes for each prediction. Tesseract is used to extract the words of each document, including the position and size of these words. For each field the overlap of the extracted words is calculated and assigned to the prediction with an overlap of over 50 percent. To receive a percentage score of text not matching the ground truth meta information per field, the sum of positive results is divided by the total occurrences.

\section{Experiments}
Recognition performance of five header fields (company-name, company-address, invoice-number, invoice-amount and invoice-date) and three line item fields (item-name, item-quantity and item-amount) are evaluated in this work.

For the experiments, ten different templates are designed to generate 12.000 documents (8000 training, 1000 validation, 3000 test) in total. To create varying documents, each field is shifted by a random offset. The font "tiro" was used during document creation to receive a constant OCR-Engine performance. All three models were trained and evaluated once with the in Chargrid proposed pre-processing method and once with the SpaCy pre-processed documents. Training took, depending on model size, between 50 to 70 epochs to fully converge with a training time of 8 hours for the smaller models, up to 16 hours for the models using all three decoders.

\section{Results}
\label{Results}
The header-field results, which can be seen in table \ref{table:Header-Fields-Results}, and the list item field results, which can be seen in table \ref{table:List-Item-Fields-Results}, show that the ground truth masks, generated in the pre-processing step, do not reach 100 percent in all cases.

\begin{table}
 \centering
 \caption{Semantic segmentation decoder: header fields results}
\begin{tabular}{ c c c c c c}
    \hline
    Experiment & company-name & invoice-number & date & amount & company-address \\
    \hline\hline
    Ground Truth Mask & 99,6\% & 94,2\% & 100\% & 99,8\% & 97,9\% \\
    Chargrid Semantic & 99,5\% & 93,9\% & 100\% & 95,1\% & 97,0\% \\
    Chargrid Box Mask & 99,3\% & 93,6\% & 100\% & 94,6\% & 97,3\% \\
    Chargrid Full model & 99,4\% & 93,9\% & 99,5\% & 94,8\% & 97,3\% \\
    SpaCygrid Semantic & 99,5\% & 93,9\% & 100\% & 95,2\% & 97,2\% \\
    SpaCygrid Box Mask & 99,3\% & 93,7\% & 100\% & 95,1\% & 97,4\% \\
    SpaCygrid Full Model & 99,0\% & 93,8\% & 99,9\% & 94,5\% & 97,5\% \\
    \hline
    \end{tabular}
\label{table:Header-Fields-Results}
\end{table}
\begin{table}
 \centering
 \caption{Semantic segmentation decoder: list item fields results}
    \begin{tabular}{ c c c c }
    \hline
    Experiment & item-names & item-quantities & item-amounts \\
    \hline\hline
    Ground Truth Mask & 95,8\% & 80,0\% & 76,3\% \\
    Chargrid Semantic & 37,1\% & 30,5\% & 31,5\% \\
    Chargrid Box Mask & 21,6\% & 22,5\% & 25,4\% \\
    Chargrid Full Model & 35,2\% & 30,5\% & 30,8\% \\
    SpaCygrid Semantic & 95,4\% & 80,0\% & 75,0\% \\
    SpaCygrid Box Mask & 95,6\% & 79,9\% & 74,8\% \\
    SpaCygrid Full Model & 95,6\% & 79,9\% & 74,9\% \\
    \hline
    \end{tabular}
\label{table:List-Item-Fields-Results}
\end{table}
This error rate can be explained due to OCR-Engine miss-classifications. Characters like "7" were often classified as "1" by Tesseract for the used font "tiro".

The highest loss in performance can be observed in the invoice-amount fields (99,8 percent to 95,2 percent) and the item-amount fields (76,3 percent to 75 percent). These fields end often with a single currency character, which does not reach a 50\% overlap with the predicted bounding box. Furthermore, the findings of BERTgrid could be confirmed, as the results show that the SpaCygrid experiments outperform the Chargrid experiments, especially for the list item fields, where the performance is improved by over 50\%. It can be seen that in most cases, where the bounding box decoder was added to the model, the results do not reach the accuracy of the experiments without this decoder. The address-fields and the item-names, which do not share a rectangular shape like the other fields, achieve better results with the box mask decoder added to the model. 

\section{Discussion}
\label{Discussion}
Since no dataset was available to compare models for key information extraction from documents, the template based document generator was used to generate a benchmark dataset with 12.000 documents using 10 dynamic templates. The experiments have shown that this dataset can be used to train, compare and evaluate different models for information extraction of invoices in PDF-format. 
The model using the semantic segmentation decoder only shows the best overall performance when trained on the generated dataset, where most fields that should be extracted follow a rectangular shape. The bounding box regression decoder can be beneficial for masks with more complex field structures. 
This research project could confirm the findings of the BERTgrid paper regarding the benefits of a NLP based pre-processing method, as the pre-processing step with SpaCy improved the Chargrid results. Especially the performance gap in the extraction of list item fields between Chargrid and the ground truth masks could be drastically reduced with a performance increasing of over 50\%.
The experiments have shown that the results for a specific field depend on the selected overlap threshold, which has to be adjusted to the fields content to achieve the best performance.
It could be observed that the OCR pre-processing step is crucial for the total model performance. The limited amount of different templates and the accompanying limited variation of the fields positions result in exceptional scores which are not expected to hold in real-world scenarios.

\section{Conclusion}
The created dataset with the document generator is a first step to create a common dataset for document processing. The presented work has shown that it can be used to train, compare and evaluate different models for information extraction of invoices in PDF format. 
Furthermore, it was shown that the use of a bounding box regression decoder is beneficial for masks following a more complex structure than a rectangle. More research with different types of templates are needed to further investigate at which point the bounding box regression decoder improves performance. The benefits of a NLP based pre-processing step could be approved, while the performance of the OCR-Engine is crucial in extracting information from PDF-documents.

\section{Future Work}
The extension of the generator with more complex and varying templates would be the next step to improve the relevance of the generator for real-world scenarios. It should be investigated weather a semantic enrichment of the text with domain ontologies can improve the results of the classification.  With more training data and resources, the effect of model components that have been neglected in this work can be evaluated as well to get a deeper understanding of the different components effect on model performance.


\end{document}